\documentclass{article}
\pdfoutput=1

\usepackage{ijcai24}
\usepackage{times}
\usepackage{soul}
\usepackage{url}
\usepackage[pagebackref,breaklinks,colorlinks,citecolor=blue]{hyperref}
\usepackage[utf8]{inputenc}
\usepackage[small]{caption}
\usepackage{graphicx}
\usepackage{amsmath}
\usepackage{amsthm}
\usepackage{booktabs}
\usepackage{algorithm}
\usepackage{algorithmic}
\usepackage[switch]{lineno}
\usepackage{amssymb}
\usepackage[normalem]{ulem}
\useunder{\uline}{\ul}{}

\title{LLMRA: Multi-modal Large Language Model based Restoration Assistant}

\author{	Xiaoyu Jin $^{1}$, Yuan shi$^{1}$, Bin Xia $^{2}$, Wenming Yang$^{1}$\\
	$^1$ Tsinghua University, $^{2}$ The Chinese University of Hong Kong \\
}

\begin{document}

\maketitle

\begin{abstract}
Multi-modal Large Language Models (MLLMs) have a significant impact on various tasks, due to their extensive knowledge and powerful perception and generation capabilities.  However, it still remains an open research problem on applying MLLMs to low-level vision tasks. In this paper, we present a simple MLLM-based Image Restoration framework to address this gap, namely Multi-modal Large Language Model based Restoration Assistant (LLMRA). We exploit the impressive capabilities of MLLMs to obtain the degradation information for universal image restoration. By employing a pretrained multi-modal large language model and a vision language model, we generate text descriptions and encode them as context embedding with degradation information for the degraded image. Through the proposed Context Enhance Module (CEM) and Degradation Context based Transformer Network (DC-former), we integrate these context embedding into the restoration network, contributing to more accurate and adjustable image restoration. Based on the dialogue with the users, our method leverages image degradation priors from MLLMs, providing low-level attributes descriptions of the input low-quality images and the restored high-quality images simultaneously. Extensive experiments demonstrate the superior performance of our LLMRA in universal image restoration tasks.

\end{abstract}

\section{Introduction}

\begin{figure*}[t]
	\centering
	\includegraphics[width = 1.0\linewidth]{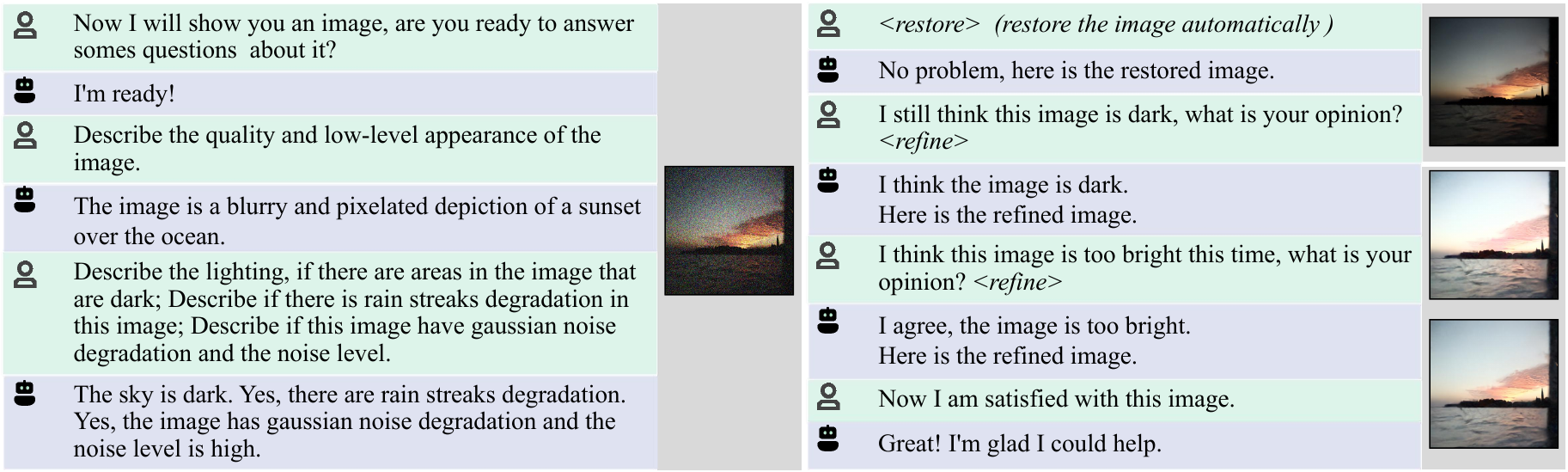}
	\caption{Example of the proposed LLMRA for universal image restoration. Based on the input image and the text input asking for the low-level attributes of the image, our method is capable of providing corresponding descriptions. Upon the \textless \textit{restore} \textgreater instruction, our LLMRA leverages the degradation descriptions from the MLLM automatically to restore the image. On the other hand, when instructed with the \textless \textit{refine}\textgreater command, LLMRA performs image restoration based on the content of the dialogue.}
\label{fig:Demo}
\vspace{-4mm}
\end{figure*}

Recently, Multi-modal Large Language Models (MLLMs), such as LLaVA~\cite{llava}, MiniGPT-4~\cite{minigpt4}, and InstructBLIP~\cite{iblip}, have garnered significant attention. Building upon the remarkable comprehension and reasoning capabilities of LLMs, MLLMs have transcended beyond the boundaries of textual inputs, harnessing their remarkable power in various domains.

However, the current exploration of MLLMs has primarily focused on high-level perception and understanding of images. The application of MLLMs only emerges in a limited range of vision-language tasks, such as image captioning~\cite{chen2015microsoft}, visual question answering~\cite{cocovqa}, and conventional computer vision tasks like segmentation~\cite{lai2023lisa} and text-to-image generation~\cite{xia2023llmga}. Recently, a benchmark called Q-bench~\cite{wu2023q} can evaluate the performance of MLLMs in low-level vision tasks, specifically in perceiving and describing low-level quality-related information using natural language. The results demonstrate that MLLMs exhibit a notable perceptual ability towards low-level visual attributes.

Image restoration is a fundamental task in the field of low-level vision, with the primary objective of recovering high-quality images from degraded counterparts. This task encompasses a diverse range of subtasks, including but not limited to image denoising, deblurring, deraining, and low-light enhancement. Presently, the existing methods predominantly concentrate on addressing specific types of image degradation, and are trained on datasets featuring only a single degradation, thereby imposing limitations on their ability to effectively restore other forms of degradation. In recent times, there has been a notable surge of interest in the task of unified image restoration. Researchers are challenged to develop a single model capable of handling images with diverse types of degradation. Several approaches have been proposed to tackle this challenge, employing techniques like degradation encoder, contrastive learning~\cite{li2022all}, and prompt learning~\cite{potlapalli2023promptir} to achieve promising results. Some approaches also leverage visual language models (VLMs) to handle a wide range of degradations~\cite{luo2023controlling}. However, when it comes to complex real-world degradations, the processing and storage capabilities of these encoders and VLMs are still limited. In particular, these methods can only directly restore images and cannot accept other instructions for restoration or optimization, which limits the application scenarios.

In this paper, we combine large-scale pretrained multi-modal large language model with image restoration networks and introduce an effective framework for universal image restoration. We refer to this novel framework as MLLM based image Restoration Assistant (LLMRA). Specifically, we utilize IDEFICS~\cite{idefics}, an open-source multi-modal language model based on Flamingo~\cite{alayrac2022flamingo}, to generate textual descriptions of the input degraded images. The text encoder of CLIP~\cite{CLIP} (a large-scale pretrained vision-language model) is employed to encode the text descriptions into text features. Using a Context Refine Module (CRM) and Context transformer, these degradation aware text features are enhanced. Finally, we incorporate them into the Degradation Context based Transformer Network (DC-former) through a Degradation Modulation Module. By effectively utilizing 
the image degradation priors obtained from the MLLMs, this framework enables the restoration network to achieve more accurate and adjustable image restoration. Our main contributions are summarised as follows:
\begin{itemize}
\item We propose a multi-modal large language model based image restoration framework, which is capable of generating restored high-quality image automatically or according to the dialogue with the users. To the best of our knowledge, LLMRA is the first work that applies MLLMs in the domain of unified image restoration.

\item To better incorporate text features into the restoration network, we propose CEM (Context Enhance Module) and DC-former (Degradation Context based Transformer Network). CEM enhances the text features and DC-former propagates the degraded information from textual features to the restoration network effectively. 

\item Our extensive experiments demonstrate the effectiveness of LLMRA, as it achieves state-of-the-art performance on various image restoration tasks, including image denoising, deraining, and low-light image enhancement.
\end{itemize}

\begin{figure*}[t]
	\centering
	\includegraphics[width = 0.99\linewidth]{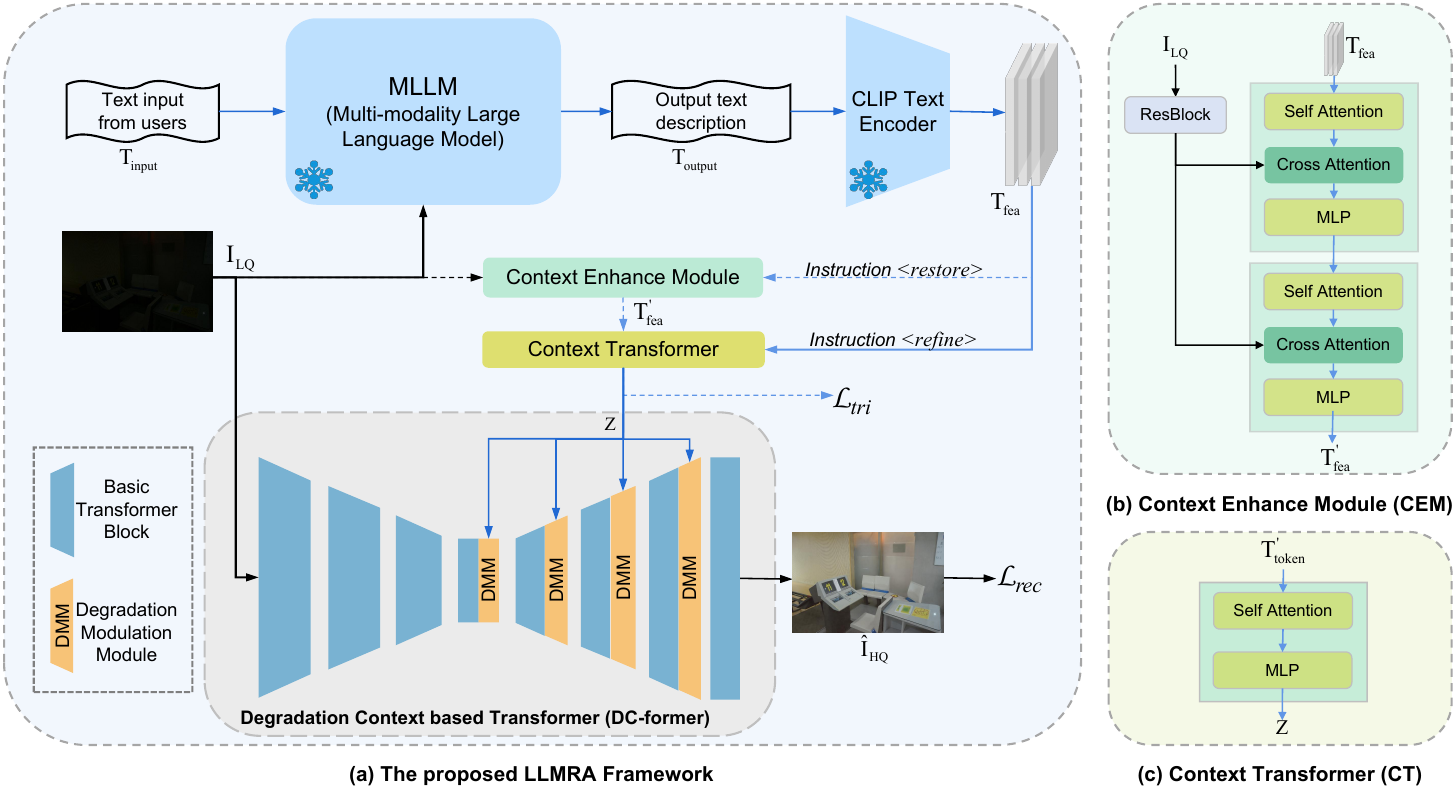}
	\caption{The overview of the proposed LLMRA.
\textbf{(a)} The proposed LLMRA Framework. DEN, CT and DC-former are used to refine and incorporate the degradation information into the restoration network.
\textbf{(b)} Context Enhance Module (CEM).
\textbf{(c)} Context Transformer (CT).
}
\label{fig:pipeline}
\end{figure*}

\section{Related Works}

\textbf{Unified Image Restoration. }
Although there has been considerable attention given to single degradation image restoration methods~\cite{zamir2022restormer,xia2023diffir}, the exploration of unified image restoration for multi-degradation remains limited. Some research has focused on addressing image degradation caused by various weather conditions such as snow, fog, and rain ~\cite{li2020all,valanarasu2022transweather}. However, these studies often train specific encoders or decoders for each weather degradation, which lacks scalability as it requires prior knowledge of specific degradation types. Li et al. proposed a unified model called AirNet ~\cite{li2022all} for denoising, deraining, and dehazing. AirNet incorporates contrastive learning to train an additional encoder, enabling implicit modeling of degradation information in the input image. These learned representations are then utilized in the main restoration network to predict the offsets of adaptable convolutions for restoration. PromptIR~\cite{potlapalli2023promptir} designed a visual prompt generation module that combines a learned degradation prompt tensor to obtain degradation features. DA-CLIP~\cite{luo2023controlling} combines a large-scale pretrained visual language model with an image restoration network and demonstrates competitive performance across the ten degradation tasks. \\
\textbf{Text-driven Image Generation. }
In recent years, there has been a rapid rise in text-based
image generation works. Several works~\cite{crowson2022vqgan,abdal2022clip2stylegan}have employed a combination of pre-trained generative models and CLIP to guide the generation process towards a desired target description. Additionally, latent diffusion model~\cite{rombach2022high} are proposed, which enables training diffusion models with limited computational resources while preserving their quality and flexibility by operating in the latent space.  In addition to these prompt-driven approaches, there have been advancements in instruction-based editing methods~\cite{geng2023instructdiffusion,brooks2023instructpix2pix} , which involve modifying a source image based on specific instructions.\\
\textbf{Multi-modal Large Language Models. }  Recent years, Large Language Models (LLMs)~\cite{llama,alpaca} have significantly contributed to conversational AI and beyond. Subsequently, attention has been directed towards advancing Multi-modal Large Language Models (MLLMs), aiming to equip LLMs with the ability to comprehend both text and images, enabling them to generate textual responses. For instance, Flamingo~\cite{alayrac2022flamingo} incorporates image encoding into the attention layer of the LLM. BLIP-2~\cite{BLIP-2} employs Q-Former to transform input images into queries. Besides, LLaVA ~\cite{llava} adopts CLIP to encode images into image embeddings, which are then concatenated with text embeddings. Then, MLLMs are adopted in various CV tasks~\cite{xia2023llmga}.

\section{Proposed Method}
In this section, we present a comprehensive description of the proposed method, which encompasses the generation of text features, the network architectures and the loss functions.

\textbf{Training. } As illustrated in Figure~\ref{fig:pipeline}, with the instruction \textless \textit{refine}\textgreater, the restoration network is first trained with accurate LQ image degradation descriptions, where the descriptions are artificially generated. Subsequently, under the \textless \textit{restore}\textgreater instruction, the Context Embedding Module (CEM) is incorporated. During this process, the textual input of degradation descriptions is provided by the MLLM. CEM is responsible for leveraging the features of the image to enhance the description generated by MLLM, thereby making it more closely aligned with the accurate depiction of degradation. For the task of unified image restoration, we consider three commonly encountered degradation types: noise, rain, and low illumination. These degradation types encompass both additive and multiplicative forms of degradation, thereby exhibiting generalization capabilities.

\textbf{Inference. }
When presented with the instruction \textless \textit{restore} \textgreater, the process initiates by taking a given degraded image $\mathbf{I}_{LQ}$ and a text prompt that solicits information regarding the degradation. These inputs are fed into the MLLM. Subsequently, the MLLM generates a descriptive text that effectively captures the low-level characteristics of the LQ image. The resulting text description is then encoded using the CLIP text encoder, yielding text feature $\mathbf{T}_{fea}$. These features are subsequently processed by the Context Enhance Module (CEM) and the Context Transformer (CT) to obtain the degradation context $\mathbf{Z}$. Finally, the context $\mathbf{Z}$ is supplied to the DC-former network for the restoration of the degraded images. When received the \textless \textit{refine}\textgreater instruction, the CEM step is omitted. The restoration takes the dialogue with the users as the text input to realize adjustable restoration.

\subsection{Generation of the Text Feature}
Figure~\ref{fig:pipeline}\textbf{(a)} illustrates the process of generating text features that contain information about image degradation in our approach. We utilized idefics-9b-instruct~\cite{idefics}, a Multi-modal Large Language Model with 9 billion parameters, as the foundation of our approach. This model is designed to process both image and text sequences as input and generate coherent text as output. 

To fully leverage the vast knowledge and amazing perceptual capabilities of MLLMs, We carefully devised instructions for text input. These instructions include three specific questions related to the mentioned degradation types (i.e., noise, rain, and low-light conditions). As depicted in Figure~\ref{fig:Demo}, the large-scale model can generate promising responses to user queries based on the image information.

Next, these output text descriptions are encoded into text features $\mathbf{T}_{fea} \in \mathbb{R}^{77 \times 512}$, using the text encoder of CLIP. The aforementioned procedure employs pretrained models, we do not need fine-tuning on them. By denoting the input text instructions and degraded image as $\mathbf{T}_{input}$ and $\mathbf{I}_{LQ}$, this process can be formulated as:
\begin{equation}
\label{eq:MLLM}
\mathbf{T}_{fea}=\mathcal{F}_{CLIP}(\mathcal{F}_{MLLM}(\mathbf{T}_{input},\mathbf{I}_{LQ})),
\end{equation}
where $\mathcal{F}_{MLLM}$ and $\mathcal{F}_{CLIP}$ indicate the text encoders of IDEFICS and CLIP, respectively.

\begin{figure}[t]
	\centering
	\includegraphics[width = 1\linewidth]{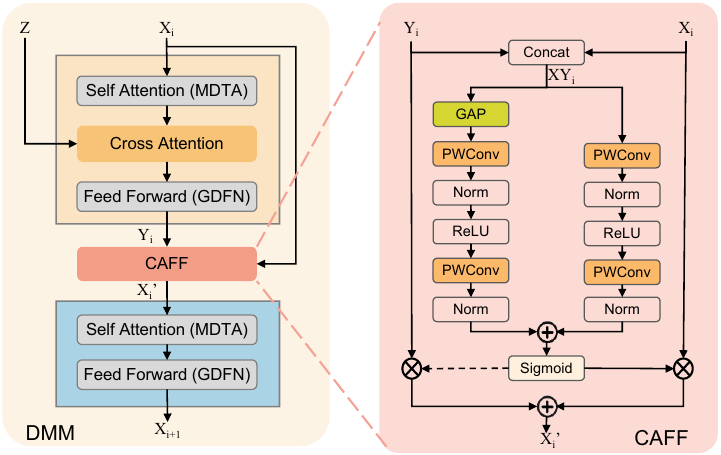}
 \vspace{-6mm}
	\caption{Degradation Modulation Module (DMM) in DC-former. }
\label{fig:DMM}
 \vspace{-4mm}
\end{figure}

\subsection{Context Enhance Module}
Under the instruction \textless \textit{refine}\textgreater training recipe, the textual descriptions are artificially generated, which is accurate and directly corresponds to the specific type of image degradation. While the instruction \textless \textit{restore} \textgreater requires the model to automatically restore the image without other priors. Due to the potential inaccuracies in the descriptions generated by MLLM, the context Enhance Module (CEM) is proposed to utilize the image features to enhance the degradation descriptions generated by MLLM. The goal is to bring these descriptions as close as possible to accurate representations of image degradation. As shown in Figure~\ref{fig:pipeline}~\textbf{(b)}, for an input LQ image $\mathbf{I}_{LQ}\in \mathbb{R}^{3 \times H \times W }$, we obtain the shallow image feature through a convolutional ResBlock. Combining the shallow image feature, we process the text features $\mathbf{T}_{fea}$ through two text cross transformers, this process is formulated as:
\begin{equation}
\label{eq:CEM}
\mathbf{T}^{\prime}_{fea}=\operatorname{CEM} (\mathbf{I}_{LQ}, \mathbf{T}_{fea})
\end{equation}
where $\mathbf{T}^{\prime}_{fea} \in \mathbb{R}^{77 \times 512}$ refers to the enhanced text features. After that, $\mathbf{T}^{\prime}_{fea}$ (or $\mathbf{T}_{fea}$) is processed by Context Transformer (CT) to get the degradation context embeddings $\mathbf{Z}$. CT is a single vanilla transformer~\cite{vaswani2017attention} consists of a self attention and multi-layer perceptron. 

As mentioned above, we need to bring $\mathbf{Z}$ as close as possible to accurate representations of image degradation. To this end, we leverage an triplet loss to learn $\mathbf{Z}$ by maximizing the consistency with the postive
inputs while minimizing the consistency between the negative ones. To be specific, for a degradation context $\mathbf{Z}$, $\mathbf{Z^+}$ and $\mathbf{Z^-}$ are the corresponding positive and negative counterpart, respectively. Then, the triplet loss $\mathcal{L}_{tri}$ could be reformulated as:
\begin{equation}
\label{eq:triplet}
\mathcal{L}_{tri} = \sum_{i=1}^{N} \left[ \left\| \mathbf{Z}_i - \mathbf{Z^+}_i \right\|_2^2 - \left\| \mathbf{Z}_i - \mathbf{Z^-}_i \right\|_2^2 + \alpha \right]_+
\end{equation}
where $\alpha$ refers to the margin of the loss.

\subsection{Degradation Context based Transformer}
With the degradation context $\mathbf{Z}$ obtained from CT, the Degradation Context based Transformer Network (DC-former) is employed to restore the high-quality image from the input with unknown degradation. The architecture of DC-former, depicted in Figure~\ref{fig:pipeline}\textbf{(a)}, consists of multiple stacked basic transformer blocks and Degradation Modulation Modules (DMM), organized in a UNet-shaped architecture. This design allows for effective information flow and contextual understanding, enabling the model to restore the image while considering the specific degradation characteristics.

As shown in Figure~\ref{fig:DMM}, each DMM consists of an image cross attention transformer (yellow box), a Concatenate Attention Feature Fusion (CAFF) module and a basic transformer block (blue box) from Restormer~\cite{zamir2022restormer}. The basic transformer block is composed of a Multi-Dconv head transposed attention (MDTA) and Gated-Dconv feedforward network (GDFN), which allow more effective feature interactions. The process is formulated as:
\begin{equation}
\label{eq:DMM}
\mathbf{X}_{i+1}= \operatorname{DMM}(\mathbf{X}_{i}, \mathbf{Z})
\end{equation}
where $\mathbf{X}_i$ and $\mathbf{X}_{i+1}$ denote the input and output feature maps.

In CAFF, we first concatenate $\mathbf{X}_i$ and $\mathbf{Y}_i$ as $\mathbf{XY}_i$. Inspired by~\cite{dai2021attentional}, the feature maps are processed with two branch to get local and global information and aggregated at the end. 
The local channel context $\mathcal{L}(\mathbf{XY}_i) \in \mathbb{R}^{C \times H \times W}$ is computed via a bottleneck structure as follows:
\begin{equation}
\begin{aligned}
\small
\mathcal{L}(\mathbf{XY}_i ) = &\operatorname{Norm} (\operatorname{PWConv}_2 ( \\
&\delta( \operatorname{Norm} (\operatorname{PWConv}_2 (\mathbf{XY}_i)))))
\end{aligned}
\end{equation}
where $\operatorname{Norm}$ refers to Layer Normalization (LN), $\operatorname{PWConv}_2$ denotes point-wise convolution (PWConv), $\delta$ denotes the Rectified Linear Unit (ReLU). Note that the kernel sizes of the two $\operatorname{PWConv}_2$ are $2C \times 2C \times 1 \times 1$ and $2C \times C \times 1 \times 1$, respectively. As a result, $\mathcal{L}(\mathbf{X}_i)$ preserves the same shape as the input feature, allowing for the preservation and emphasis of intricate details in the low-level features.

In the global branch, the features are first processed through a global average pooling (GAP), followed by similar operations as $\operatorname{PWConv}_1$, LN, ReLU, $\operatorname{PWConv}_1$ and LN, finally get the global channel context $\mathcal{G}(\mathbf{XY}_i)$. The $\operatorname{PWConv}_1$ here is for one dimension. It is formulated as:
\begin{equation}
\begin{aligned}
\mathcal{G}(\mathbf{XY}_i) = &\operatorname{Norm} (\operatorname{PWConv}_1 ( \\
&\delta( \operatorname{Norm} (\operatorname{PWConv}_1 (\operatorname{GAP}(\mathbf{XY}_i)) ))))
\end{aligned}
\end{equation}
By incorporating the global channel context $\mathcal{G}(\mathbf{XY}_i)$ and local channel context $\mathcal{L}(\mathbf{XY}_i)$ the modulated feature $\mathbf{X}^{\prime}_i$ can be obtained as follows:
\begin{equation}
\mathbf{X}^{\prime}_i = \mathbf{X}_i \otimes \mathbf{W}(\mathbf{XY}_i) + (1- \mathbf{W}(\mathbf{XY}_i)) \otimes \mathbf{Y}_i
\end{equation}
\begin{equation}
 \mathbf{W}(\mathbf{XY}_i) = \sigma ((\mathcal{L}(\mathbf{XY}_i) \oplus \mathcal{G}(\mathbf{XY}_i) ),
\end{equation}
where $\sigma$ denotes Sigmoid operation, $\mathbf{W}$ denotes the attention weights. $\oplus$ denotes the broadcasting addition and $\otimes$ denotes the element-wise multiplication.

\begin{figure*}[ht!]
	\centering
	\includegraphics[width = 1.0\linewidth]{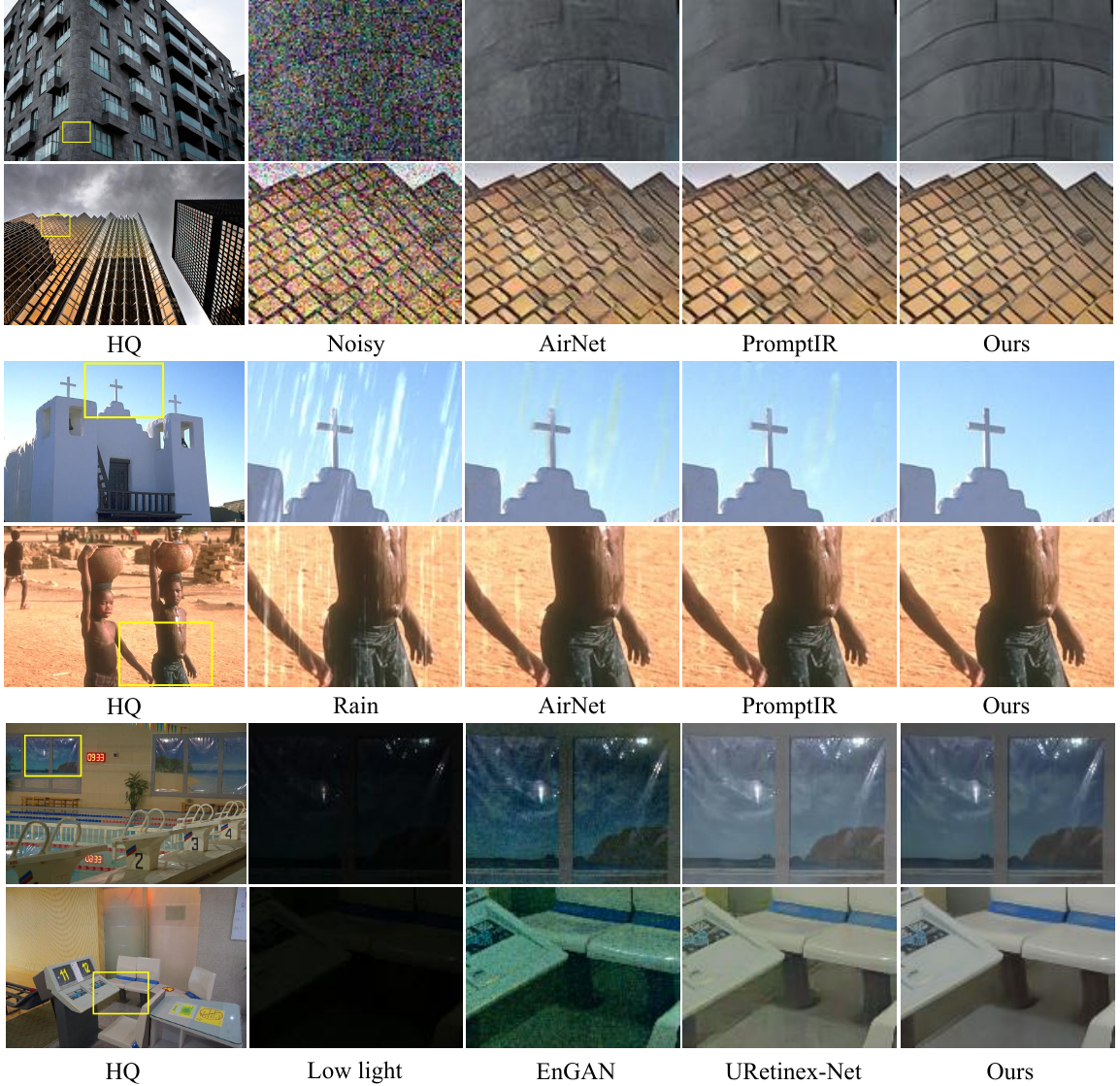}
	\caption{Visual comparisons with the SOTA methods. Rows 1-2, 3-4, 5-6 rows display the results of image denoising, image deraining and low light image enhancement, respectively. The test images are from Urban100, Rain100L and LOLv1. Zoom in for better visualization.}
\label{fig:Visual_Comparism}
\end{figure*}

\subsection{The Objective Function}
As mentioned above, when training the models using the \textless \textit{restore}\textgreater and \textless \textit{refine}\textgreater instructions, we employ distinct objective functions to optimize the process.
\begin{equation}\label{eq:loss_refine}
    \mathcal{L}_{refine} = \mathcal{L}_{rec}
\end{equation}
\begin{equation}\label{eq:loss_restore}
    \mathcal{L}_{restore} = \mathcal{L}_{rec} + \mathcal{L}_{tri}
\end{equation}
where $\mathcal{L}_{tri}$ refers to the triplet loss (equation~\ref{eq:triplet}) and $\mathcal{L}_{rec} = \left\|\mathbf{I}_{HQ} - \mathbf{\hat{I}}_{HQ}\right\|_1$ represents the reconstruction loss, which caculates the L1 norm between the ground truth $\mathbf{I}_{HQ}$ and the recovered high quality image $\mathbf{\hat{I}}_{HQ}$. 

\section{Experiments}

\subsection{Experimental Settings}
To demonstrate the effectiveness of the proposed LLMRA, we perform the evaluation on three representative image restoration tasks: image denoising, image deraining, and low light image enhancement. We train a unified model that can recover images across all three degradation types.

\begin{table*}[ht!]
\centering
\caption{Denoising comparisons in the single-task setting on BSD68 and Urban100 datasets. Top rows: methods under the single-task setting. Bottom rows: methods under the all-in-one setting. The optimal and sub-optimal PSNR/SSIM$\uparrow$ results are highlighted using bold and underlined, respectively.}
\vspace{-1mm}
\label{tab:denoising}
\begin{tabular}{lccc|ccc} 
\hline
\multicolumn{1}{c|}{}          & \multicolumn{3}{c|}{BSD68}                                                                                         & \multicolumn{3}{c}{Urban100}                                                                                      \\
\multicolumn{1}{l|}{Method}   & $\sigma = 15$                             & \multicolumn{1}{c}{$\sigma = 25$} & \multicolumn{1}{c|}{$\sigma = 50$} & $\sigma = 15$                             & \multicolumn{1}{c}{$\sigma = 25$} & \multicolumn{1}{c}{$\sigma = 50$} \\ \hline
\multicolumn{1}{l|}{DnCNN}    & 33.89/0.9290                              & \multicolumn{1}{c}{31.23/0.8830}  & \multicolumn{1}{c|}{27.92/0.7896}  & 32.98/0.9314                              & \multicolumn{1}{c}{30.81/0.9015}  & \multicolumn{1}{c}{27.59/0.8331}  \\
\multicolumn{1}{l|}{IRCNN}    & 33.87/0.9285                              & \multicolumn{1}{c}{31.18/0.8824}  & \multicolumn{1}{c|}{27.88/0.7898}  & 27.59/0.8331                              & \multicolumn{1}{c}{31.20/0.9088}  & \multicolumn{1}{c}{27.70/0.8396}  \\
\multicolumn{1}{l|}{FFDNet}   & 33.87/0.9290                              & \multicolumn{1}{c}{31.21/0.8821}  & \multicolumn{1}{c|}{27.96/0.7887}  & 33.83/0.9418                              & \multicolumn{1}{c}{31.40/0.9120}  & \multicolumn{1}{c}{28.05/0.8476}  \\ \hline
\multicolumn{1}{l|}{AirNet}   & \multicolumn{1}{l}{33.85/0.9293}          & 31.22/0.8837                      & 27.98/0.7933                       & \multicolumn{1}{l}{33.89/0.9419}          & 31.52/0.9137                      & 28.19/0.8520                       \\
\multicolumn{1}{l|}{DA-CLIP}  & \multicolumn{1}{l}{26.34/0.6821}          & 25.77/0.6531                      & 24.31/0.5712                       & \multicolumn{1}{c}{-}                     & -                                 & -                                 \\
\multicolumn{1}{l|}{PromptIR} & \multicolumn{1}{l}{{\ul 33.91/0.9296}}    & {\ul 31.28/0.8840}                & {\ul 28.03/0.7926}                 & \multicolumn{1}{l}{{\ul 33.93/0.9417}}    & {\ul 31.52/0.9121}                & {\ul 28.17/0.8498}                \\
\multicolumn{1}{l|}{Ours}  & \multicolumn{1}{l}{\textbf{34.01/0.9302}} & \textbf{31.37/0.8849}             & \textbf{28.13/0.7930}               & \multicolumn{1}{l}{\textbf{34.12/0.9435}} & \textbf{31.79/0.9163}             & \textbf{28.56/0.8578}             \\ \hline
\end{tabular}
\end{table*}

\begin{table*}[ht!]
  \centering
  \caption{Deraining results on Rain100L. Left columns: methods under single-task setting. Right columns: methods under all-in-one setting. The optimal and sub-optimal results are highlighted using bold and underlined, respectively.}
  \label{tab:deraining}
\begin{tabular}{c|cccc|cccc}
\hline
     & UMR   & SIRR  & MSPFN & Restormer   & AirNet & DA-CLIP & PromptIR    & Ours        \\ \hline
PSNR$\uparrow$ & 32.39 & 32.37 & 33.50 & {\ul 37.57} & 34.90  & 35.19   & 37.32       & \textbf{38.93} \\
SSIM$\uparrow$ & 0.921 & 0.926 & 0.948 & 0.974       & 0.968  & 0.960   & {\ul 0.979} & \textbf{0.984} \\ \hline
\end{tabular}   
\end{table*}

\begin{table*}[ht!]
  \centering
  \caption{Low light image enhancement results on LOL-v1. Left columns: methods under single-task setting. Right columns: methods under all-in-one setting. The optimal and sub-optimal results are highlighted using bold and underlined, respectively. }
  \label{tab:lowlight}
  \vspace{-1mm}
\begin{tabular}{c|cccccc|cc}
\hline
     & Retinex-Net & UFormer & EnGAN & KinD  & URetinex-Net         & Restormer & DA-CLIP        & Ours        \\ \hline
PSNR$\uparrow$ & 16.40   & 16.36   & 17.56 & 20.86 & 21.33                & 22.43     & \textbf{ 23.40} & {\ul 23.30}    \\
SSIM$\uparrow$ & 0.500   & 0.771   & 0.665 & 0.790 & {\ul{0.834}} & 0.823     & 0.811          & \textbf{0.846} \\ \hline
\end{tabular} 
\end{table*}

\textbf{Implementation Details.} 
The architecture of the DC-former consists of a 4-level encoder-decoder, with varying numbers of Transformer blocks at each level, specifically [4, 6, 6, 8] from level-1 to level-4. We employ one DMM between every two consecutive decoder levels, totaling 4 DMMs in the overall DC-former network. The channel size of DC-former is set to 48. The model is trained with a batch size of 4. The network is optimized with Adam optimizer ($\beta_1 = 0.9, \beta_2 = 0.999$) with learning rate $1e^{-4}$ for 800k iters. During training, we utilize cropped patches of size 128 x 128 as input, and to augment the training data, random horizontal and vertical flips are applied to the input images.

\textbf{Datasets. }
In our experiments, we prepare several datasets for the training of these three tasks. For image denoising, we use WED~\cite{ma2016waterloo} for training, which contains 4744 images. Testing is performed on BSD68~\cite{martin2001database} and Urban100~\cite{huang2015single} datasets. From clean images of WED BSD68 and Urban100, we generate the noisy images by adding Gaussian noise with different noise levels $\sigma \in \{15,25,50\}$.  
For image deraining, we use the data from~\cite{yang2019joint}, including 1800 paired light rainy images for training and 100 images for testing. 
For low light image enhancement, we use LOL-v1 dataset ~\cite{wei2018deep}, including 485 low/normal light images pairs for training and another 15 images for testing.

\subsection{Comparison with State-of-the-Art Approaches}
For comparing with the SOTA approaches, we trained the proposed LLMRA in all-in-one settings by optimizing the network (without CEM) with $\mathcal{L}_{refine}$ (equation~\ref{eq:loss_refine}).
We compare our LLMRA with several unified image restoration approaches as well as specific degradation restoration methods on three tasks. More precisely, we compare DnCNN~\cite{zhang2017beyond}, IRCNN~\cite{zhang2017learning}, FFDNet~\cite{zhang2018ffdnet}, AirNet~\cite{li2022all}, PromptIR~\cite{potlapalli2023promptir} and DA-CLIP~\cite{luo2023controlling} for image denoisig. 
We compare UMR~\cite{yasarla2019uncertainty}, SIRR~\cite{wei2019semi}, MSPFN~\cite{jiang2020multi}, Restormer~\cite{zamir2022restormer}, AirNet~\cite{li2022all}, PromptIR~\cite{potlapalli2023promptir}, and DA-CLIP~\cite{luo2023controlling} for image deraining.
We compare Retinex~\cite{wei2018deep}, UFormer~\cite{Wang_2022_CVPR}, EnGAN~\cite{jiang2021enlightengan}, KinD~\cite{zhang2019kindling}
URetinex-Net~\cite{wu2022uretinex}, Restormer~\cite{zamir2022restormer} and DA-CLIP~\cite{luo2023controlling} for low light image enhancement.

\textbf{Quantitative Comparison.  }
Table~\ref{tab:denoising} presents results of image denoising. It shows that our LLMRA achieves 0.39dB for PSNR improvement over PromptIR for noise level $\sigma=50$ on Urban100 dataset. Similar trends can be observed for deraining tasks. On the deraining task (Table~\ref{tab:deraining}), our method yields performance gains of 1.61 dB over PropmtIR.  For low light image enhancement , our LLMRA achieves 0.035 for SSIM improvement over DA-CLIP. Our method even outperforms the restormer 
for image deraining and low-light image enhancement, which is trained in the single-task settings.

\textbf{Qualitative Comparison.  }
In addition, we provide visual examples to illustrate the effectiveness of our proposed method. Figure~\ref{fig:Visual_Comparism} showcases the results of the three tasks. For image denoising, our LLMRA outperforms other state-of-the-art methods by effectively removing noise from the image without excessively blurring it. Similarly, the middle rows demonstrate the efficacy of our approach in the deraining task, as it successfully eliminates rain streaks and produces rain-free images. For low light image enhancement, previous methods often suffered from issues such as color distortion, over/underexposed regions, or failure to suppress noise in specific areas. In contrast, our approach excels in enhancing visibility, reliably enhancing the image without introducing artifacts, and robustly preserving the natural color.

\subsection{Impact of the Text Inputs}
\begin{figure}[t]
	\centering
	\includegraphics[width = 1.0\linewidth]{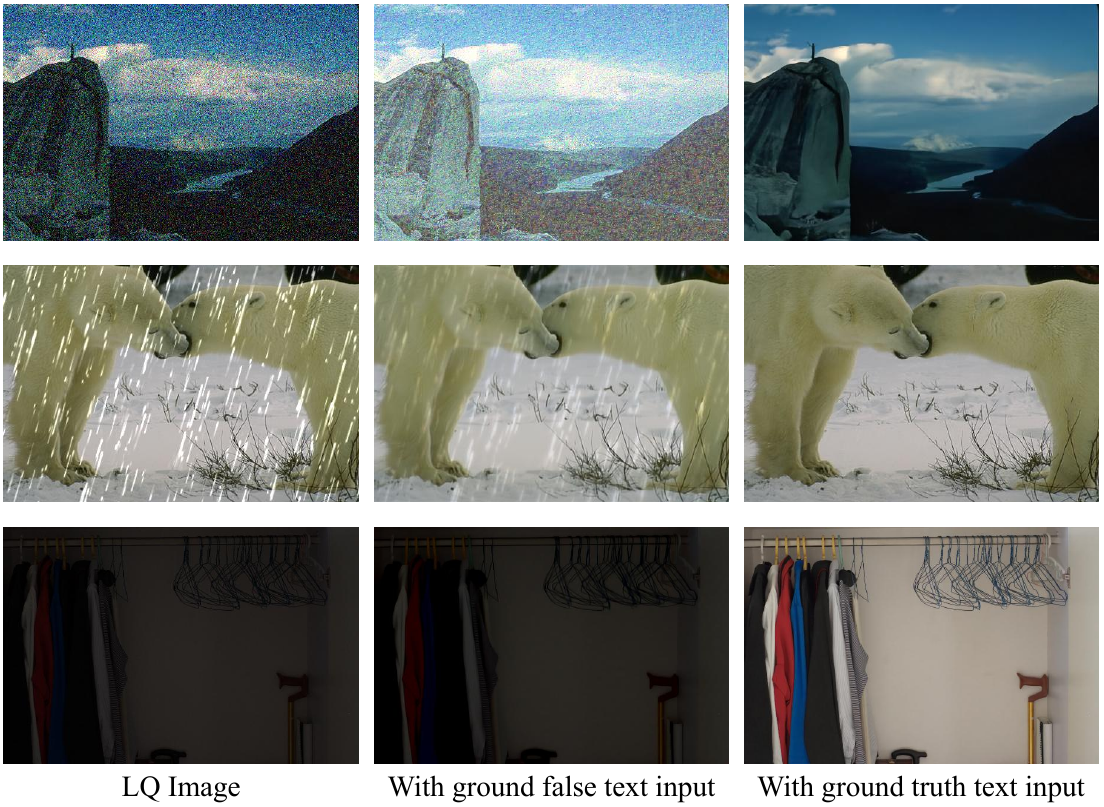}
	\caption{Impact of the text input. }
\label{fig:Text_impact}
\end{figure}

We manage to use text information to assist image restoration, as text input is more readily available and allows for adjustable and interactive restoration manner through dialogue with the MLLMs. To verify the impact of the text inputs, we prepared two set of text descriptions for the test datasets, which are called ``ground truth'' and ``ground false'' text input. As shown in Figure~\ref{fig:Text_impact}, the task is image denoising for the first row (14037.png from BSD68 with $\sigma=50$), the ground truth text description could be ``The image is well lit. No rain streaks detected. The image has gaussian noise degradation and the noise level is high.'' Conversely, the ground false text description is would be completely opposite, like ``The image is dark. The image is degraded by rain streaks. No noise detected.'' From Figure~\ref{fig:Text_impact}, it is evident that the presence of ground truth text input results in effective noise removal without any other modifications. Conversely, when ground false text input is used, the noise persists but the lighting is enhanced. Similar clue could also be drawn from the quantitative results in Table~\ref{tab:Text_impact}, when confronted with accurate and erroneous textual input, the disparity in the results of restoration is substantial.

\subsection{Ablation study}

\begin{table}[t]
  \centering
  \caption{Quantitative results for the impact of the text input, evaluated on BSD68 ($\sigma=50$), Rain100L and LOLv1 dataset. ``with gt text'' means input ground truth text descriptions. ``with gf text'' means input ground false text descriptions. }
  \label{tab:Text_impact}
    \begin{tabular}{c|cc}
    \hline
             & with gf text                     & with gt text                     \\ \hline
    BSD68    & \multicolumn{1}{l}{14.46/0.4790} & \multicolumn{1}{l}{28.13/0.7930} \\
    Rain100L & 20.11/0.8302                     & 38.93/0.9842                     \\
    LoLv1    & 7.59/0.1440                      & 23.30/0.8457                     \\ \hline
    \end{tabular}
\end{table}

\begin{table}[t]
  \centering
  \caption{Ablation study on the impact of CEM.  Results are reported on BSD68 ($\sigma=50$), Rain100L and LOLv1 datasets. The best results are shown in boldface.} 
  \label{tab:Z}
\begin{tabular}{c|cc}
\hline
         & w.o. CEM                         & Ours                    \\ \hline
BSD68    & 25.18/0.6913                     & \textbf{28.11/0.7964} \\
Rain100L & 26.54/0.8838                     & \textbf{38.64/0.9831}                     \\
LoLv1    & 17.51/0.6999                     & \textbf{20.19/0.8243}                     \\ \hline
\end{tabular}
\end{table}
\begin{table}[t]
  \centering
  \caption{Ablation study on the way of modulating the text features. Results are reported on BSD68 ($\sigma=50$), Rain100L and LOLv1 datasets. The best results are shown in boldface.}
  \label{tab:DMM}
\begin{tabular}{c|cc}
\hline
         & w.o. DMM                        & Ours                                      \\ \hline
BSD68    & \multicolumn{1}{l}{28.02/0.7913} & \multicolumn{1}{l}{\textbf{28.13/0.7930}} \\
Rain100L & 37.71/0.9796                     & \textbf{38.93/0.9842}                     \\
LoLv1    & 19.40/0.8013                     & \textbf{23.30/0.8457}                     \\ \hline
\end{tabular}
\end{table}

\textbf{Impact of CEM.  } 
To verify the impact of Context Enhance Module (CEM) on enhancing the text descriptions obtained from the MLLM in the universal image restoration task, we carry out some experiments.  In this section, a set of predefined specific questions related to the mentioned degradation types (i.e., noise, rain, and low-light conditions) are sent to the MLLM, and it would generate corresponding responses to be the text descriptions for further guiding the restoration. We restore the images with and without CEM under these conditions. The results are shown in Table 2, revealing a significant improvement in the restoration outcomes when CEM is incorporated.

\textbf{The way of modulating the text features.  } 
In the domain of text-to-image generation~\cite{rombach2022high}, researchers commonly employ a denoising UNet with a cross transformer as the basic module to modulate the text features. However, in our proposed method, we utilize DMMs for the degradation context modulation. In order to validate the effectiveness of our method, we follow the approach of these text-to-image generation methods by removing the CAFF modules and stacking the cross transformers in the decoder. The experimental results are presented in Table~\ref{tab:DMM}, which demonstrates the effectiveness of the proposed DMM.

\section{Conclusion}

This paper introduces LLMRA, a novel framework that leverages multi-modal large language models for universal image restoration. The core contribution of our framework is utilizing the MLLM and a text-guided restoration network to realize a more accurate, adjustable and interactive restoration manner. The Context Enhance Module and the Degradation Context based Transformer Network are proposed to effectively enhance the degradation information and incorporate it into the restoration network. Experimental evaluation on unified image restoration tasks demonstrates that LLMRA leads to significant performance on image denoising, image deraining, and low light image enhancement. 
Nevertheless, it is important to acknowledge some limitations of the proposed LLMRA. The performance of LLMRA may fluctuate with the performance of MLLM, as it may provide uninformative or even harmful answers of the degradation information, thus affecting the quality of restoration. Fortunately, users can \textless \textit{refine}\textgreater the results by engaging in subsequent dialogue. More ever, our experiments are currently limited to only three tasks. Although these three tasks are representative to some extent, as they encompass both additive and multiplicative degradation. In future research, we aim to broaden the scope of our investigation to encompass a wider range of restoration tasks involving different types of degradation.
\appendix

\bibliographystyle{named}
\bibliography{main}

\end{document}